\def\year{2020}\relax
\documentclass[letterpaper]{article} 
\usepackage{aaai20}  
\usepackage{times}  
\usepackage{helvet} 
\usepackage{courier}  
\usepackage[hyphens]{url}  
\usepackage{graphicx} 
\usepackage{graphicx}  
\title{A Joint Model for Definition Extraction \\ with Syntactic Connection and Semantic Consistency}

\author{
Amir Pouran Ben Veyseh\textsuperscript{\rm 1}\thanks{This work was done when the first author was an intern at Adobe Research.}, Franck Dernoncourt\textsuperscript{\rm 2}, Dejing Dou\textsuperscript{\rm 1} and Thien Huu Nguyen\textsuperscript{\rm 1,3} \\
\textsuperscript{\rm 1}Department of Computer and Information Science, University of Oregon, Eugene, Oregon 97403, USA\\
\textsuperscript{\rm 2}Adobe Research, San Jose, CA, USA\\
\textsuperscript{\rm 3}VinAI Research, Hanoi, Vietnam\\
{\tt \{apouranb, dou, thien\}@cs.uoregon.edu} \\
{\tt dernonco@adobe.com}
}

\begin{document}

\maketitle

\begin{abstract}
Definition Extraction (DE) is one of the well-known topics in Information Extraction that aims to identify terms and their corresponding definitions in unstructured texts. This task can be formalized either as a sentence classification task (i.e., containing term-definition pairs or not) or a sequential labeling task (i.e., identifying the boundaries of the terms and definitions). The previous works for DE have only focused on one of the two approaches, failing to model the inter-dependencies between the two tasks. In this work, we propose a novel model for DE that simultaneously performs the two tasks in a single framework to benefit from their inter-dependencies. Our model features deep learning architectures to exploit the global structures of the input sentences as well as the semantic consistencies between the terms and the definitions, thereby improving the quality of the representation vectors for DE. Besides the joint inference between sentence classification and sequential labeling, the proposed model is fundamentally different from the prior work for DE in that the prior work has only employed the local structures of the input sentences (i.e., word-to-word relations), and not yet considered the semantic consistencies between terms and definitions. In order to implement these novel ideas, our model presents a multi-task learning framework that employs graph convolutional neural networks and predicts the dependency paths between the terms and the definitions. We also seek to enforce the consistency between the representations of the terms and definitions both globally (i.e., increasing semantic consistency between the representations of the entire sentences and the terms/definitions) and locally (i.e., promoting the similarity between the representations of the terms and the definitions). The extensive experiments on three benchmark datasets demonstrate the effectiveness of our approach.\footnote{Source code is available at \url{https://github.com/amirveyseh/definition_extraction}.}
\end{abstract}

\section{Introduction}

One of the cornerstones of human knowledge that can convey meaningful, concise and informative understanding of different concepts are definitions. Due to their importance, resources such as dictionaries and glossaries are created for both general and specific domains. However, manually creating these resources from raw text is expensive, time-consuming and requiring extensive domain and linguistic knowledge. Thus, automatic definition extraction from raw text has gained much attention from the computational linguistic community, finding its applications on many downstream tasks (e.g., Question Answering, Knowledge Base Completion, and Taxonomy Learning).

Formally, the task of Definition Extraction (DE) aims to detect term-definition pairs in raw text. In a coarse-grained setting, the goal is to classify the input sentences as definitional or not (i.e., sequence classification) where a sentence is definitional if it contains a term-definition pair. On the other hand, in a fine-grained setting, DE aims to recognize the terms and definitions within the sentences, usually casted as a sequence labeling task. For instance, the sentence ``\textit{The phrase \textless Term\textgreater {\bf atoms and molecules}\textless \slash Term\textgreater is explained in the dictionary by the expression of \textless Definition\textgreater {\bf building blocks of materials}\textless \slash Definition\textgreater}'' is definitional as its term and definition are specified by the corresponding tags. 

The prior work has only formulated DE as either a sequence classification or sequence labeling task, ignoring the inter-dependencies between these two tasks. This is not desirable as there exist the mutual dependencies between the two tasks for which the knowledge for one task can help to improve the performance for the other. In particular, knowing the word-level labels of the words for terms and definitions (i.e., sequence labeling) would be sufficient to infer the sentence level labels (i.e., sentence classification) as the latter task is a simplified version of the former. On the other hand, predicting the sentence level label can also improve the knowledge of the models about the different templates/patterns to express/connect terms and definitions in text that would be helpful to recognize the boundaries of the terms and definitions. Consequently, in this paper, we propose to simultaneously predict the sentence labels and word-level labels for terms and definitions in the sentences based on deep learning to benefit from the inter-dependencies between the two tasks. To the best of our knowledge, this is the first work to jointly perform sequence classification and sequence labeling for DE in the literature.



The recent works have shown that structure information available in dependency parse trees can help to improve the performance of deep learning models for DE due to their ability to capture the head-modifier relations between words in the input sentences \cite{anke:2018}. However, despite such performance gain, the application of the dependency trees in the previous work for DE is only limited to the local structures (i.e., the word-by-word relations of the trees), ignoring the global structure of the input sentences (i.e., whole dependency trees). Such global structures of the sentences are useful for DE as they can identify the important context words for the terms and definitions in the sentences (e.g., via the dependency paths between terms and definitions). These contextual information (e.g., the words ``{\it explained}'', ``{\it by}'', and ``{\it expression}'' in the above example) would help to provide better evidences to distinguish definitional sentences (i.e., for the sequence classification task) and recognize the boundaries for the terms and definitions (i.e., for the sequence labeling task). In order to address the local structure issue in the previous work for DE, in this work, we propose to explicitly model the whole structures of the sentences to better exploit the dependency trees for DE. To this end, we employ Graph Convolutional Neural Network (GCN) to obtain the structure-aware representations for the words and the input sentences based on the whole dependency trees to perform DE in this work. Afterwards, in a multi-task learning framework, in addition to the DE tasks, we seek to predict which words in the sentences belong to the dependency paths between the terms and the definitions during the training process of the models.

Another issue of the previous work on DE is that they fail to exploit the semantic consistency between the terms, the definitions and the entire sentences containing such elements. In particular, in the sentences that define terms and definitions, the definitions are in general the detailed explanations or representation of the terms while the combinations of the terms and definitions would constitute the major information of the entire sentences needed for the DE tasks. Consequently, we expect that if a deep learning model can identify the terms and definitions well, it should be able to achieve some level of similarity or consistency between the representations for the terms, the definitions and the entire sentences. In order to exploit this intuition, in this work, we propose to introduce explicit constraints to promote the similarity of such representations into the overall loss function of the models for DE. Specifically, we consider both the local and global semantic consistencies in this work.

First, in the local semantic consistency, it is a natural intuition that the semantics of the terms and their definitions are similar to each other. In this work, we argue that such intuition can be used as an inductive bias to improve the representations learned by deep learning for DE. To this end, we propose to extensively employ this intuition by introducing both direct and indirect constraints to promote the similarity/consistency between the representations of the terms and definitions. In the direct constraint, we seek to increase the dot product of the term and definition representations in the overall loss function. The indirect constraint, on the other hand, achieves the semantic relatedness between the terms and the definitions by ensuring that the representations of the definitions are more similar to the representations of the words in the terms than those for the other words in the sentences. This is implemented via a discriminator that learns to predict the similarity scores between the representation vectors of the definitions and the other words in the sentences.





Second, for the global semantic consistency, we learn the representations for the entire sentences and the term-definition pairs that are then constrained to be similar to each other. The rationale is the sentences might involve some words and/or phrases that are not directly related to the definitions, the terms and their connections (i.e., the important context words between terms and definitions). The typical representations of the entire sentences might thus incur some noisy information for DE due to the modeling of such irrelevant words/phrases. The consistency constraints with the term-definition pairs would help to restrict the topical information in the sentence representations to mainly focus on the definitions and terms for DE. However, as the sentence representations might need to additionally involve the information about the important context words between the terms and the definitions, we only attempt to impose a mild and indirect consistency constraint for the sentence and term-definition representations in this case. In particular, instead of directly enforcing the representation similarity, we only seek to ensure that the representations for the sentences can be used to predict the same latent labels as those predicted by the representations for the term-definition pairs.

Our extensive experiments in multiple DE benchmark datasets prove the effectiveness of our approach, yielding the state-of-the-art performance for such datasets over both the sequence classification and sequence labeling settings. In summary, the contributions of this work include: (i) a new joint model to perform sequence labeling and sequence classification for DE, (ii) a novel method to exploit the global syntactic structures for DE, (iii) multiple consistency enforcement techniques to achieve the local and global semantic consistency for terms, definitions, and sentences, and (iv) the state-of-the-art performance on multiple DE datasets.


\section{Related Work}


We can categorize the previous work on DE into three categories: 1) the rule-based approach: the first attempts in DE have defined linguistic rules and templates to capture patterns to express the term-definition relations \cite{JL:2001,cui:2004,cui:2005,fahmi:2006}. While the rule-based approach is intuitive and has high precision, it suffers from the low recall issue; 2) the feature engineering approach: this approach address the low recall issue by relying on the statistical machine learning models (i.e., SVM and CRF) with carefully designed features (i.e., syntax and semantics) to solve DE \cite{jin:2013,westerhout:2009}. However, this approach cannot be adapted to new domains efficiently as the designed features might be unavailable or less effective in the new domains; 3) the deep learning approach: similar to many natural language processing (NLP) tasks, deep learning has been recently shown as the state-of-the-art approach for DE due to its ability to effectively exploit the word embeddings via multiple layers of neural networks. In a recent work, \cite{anke:2018} present a deep learning model for coarse-grained DE that leverages the representation learning capacity from both convolutional neural networks (CNN) and Long-short term memory networks (LSTM). The head-modifier relations of the dependency trees are employed in this work to improve the DE performance.

The DE task is also related to the popular tasks of named entity recognition (NER) (i.e., both need to solve a sequence labeling problem) \cite{Ando:05,Ratinov:09,Florian:03,Chiu:16,Nguyen:16b} and relation extraction (RE) (i.e., both need to consider the relations/connections between two mentions in text) \cite{zhou:05,zeng:14,Nguyen:15a,Nguyen:15c,Miwa:16,Nguyen:16d} in NLP. However, DE is different from NER in that the terms and definitions in DE are in general longer and involve richer semantic structures than the names and mentions in NER. Also, in DE, the terms and definitions are usually not known in advance while the two entity mentions in RE are often given, causing the essential difference for such problems.



\section{Model}
\subsubsection{Problem Definition}
Formally, the problem of Definition Extraction (DE) is described as follows. First, for the sequence labeling task, given an input sentence/sequence $W = w_1,w_2,\ldots,w_N$ ($N$  is the number of words in the sentence and $w_i$ is the $i$-th word/token in the sentence), we need to assign a label $l_i$ to each word $w_i$ in the sentence so the resulting label sequence $L = l_1,l_2,\ldots,l_N$ would reveal the terms and definitions $W$ (i.e., word-level prediction). In this work, we use the BIO tagging schema to encode the label $l_i$ that basically defines the five possible word labels for the terms and definitions, i.e.: \textbf{B-Term}, \textbf{I-Term}, \textbf{B-Definition}, \textbf{I-Definition}, and \textbf{O} (Others). Second, for the sequence classification task, we need to make a binary decision $l$ for the overall sentence to determine if the sentence contains any term-definition pair or not (i.e., sentence-level prediction).

The model in this work is trained in a multi-task learning framework where the word-level predictions (i.e., assigning labels for all the words in the sentence for sequence labeling) and the sentence-level prediction are done simultaneously. The major modules in the proposed model include: sentence encoding, sequence labeling, sequence classification, syntactic connection, and semantic consistency.



\subsubsection{Sentence Encoding}

In order to prepare the input sentence $W$ for the following neural computation, we first transform each word $w_i \in W$ into a real-valued vector. In this work, we use the concatenation of the pre-trained word embedding of $w_i$ and the embedding vector for its part of speech (POS) in the sentence (i.e., POS embedding) to constitute the vector $e_i$ to represent $w_i$. Both the POS embeddings (initialized randomly) and the pre-trained word embeddings would be optimized during training in this work. This would convert the input sentence $W$ into a sequence of representation vector $E = e_1, e_2, \ldots, e_N$.




In the next step, to enrich the word representations with the contextualized information in $W$, we feed $E$ into a bidirectional LSTM (BiLSTM) network, producing a hidden vector sequence $H = h_1, h_2, \ldots, h_N$ as the output. Each vector $h_i$ is the concatenation of the hidden vectors from the forward and backward LSTM networks at position $i$ to capture the contextualized information for $w_i$.

Due to the sequential order to process the words in the sentence of the BiLSTM layer, a hidden vector $h_i \in H$ for $w_i$ would tend to encode the context information of the neighboring words of $w_i$ (i.e., with closer distances) more intensively than those for the other words (i.e., with farther distances). This is not desirable as the important context words to reveal the underlying semantics of $w_i$ are not necessary its neighbors and might be distributed at farther positions in the sentence. It is thus desirable to identify such important context words for each word $w_i \in W$ to compute more effective representation vectors for the words. To this end, we propose to use the whole dependency tree of $W$ as a way to link the words in the sentence to their important context words. A GCN layer \cite{kipf2016semi,Xu:18} is then applied over this dependency tree structure to enrich the representation vectors for the words in the sentence with the information from the important context words.


In particular, the GCN module involves multiple layers where each layer consumes a sequence of hidden vectors and return another hidden vector sequence as the output. Let $\hat{H}^t = \hat{h}^t_1, \hat{h}^t_2, \ldots, \hat{h}^t_N$ be the input vector sequences for the $t$-th GCN layer. The output vector sequence $\hat{H}^{t+1} = \hat{h}^{t+1}_1, \hat{h}^{t+1}_2, \ldots, \hat{h}^{t+1}_N$ of the $t$-th layer is then computed by:
 $   \hat{h}^{t+1}_i  = ReLU(W_t \bar{h}^{t+1}_i), \bar{h}^{t+1}_i  = \Sigma_{j\in N(i)} \hat{h}^t_j / deg(i)$.
 Here, $N(i)$ is index set of the neighbors of $w_i$ (including $i$ itself), $W_t$ is the weight matrix for $t$-th layer, and $deg(i)$ is the degree of $w_i$ in the dependency tree. The biases in the equations are omitted for brevity in this work.

In this work, we employ two layers for the GCN module (i.e., fine-tuned on the development datasets). The input for the first GCN layer is the sequence of hidden vectors $H = h_1,h_2,\ldots,h_N$ from BiLSTM while the output vector sequence of the last GCN layer (i.e., the second layer) is denoted by $\hat{H} = \hat{h}_1, \hat{h}_2, \ldots, \hat{h}_N$ for convenience. The representation vector in $\hat{h}_i \in \hat{H}$ would encode rich contextualized context information for $w_i$ that is augmented with the dependency structure for the important context words in $W$.

\subsubsection{Sequence Labeling}


The goal of the sequence labeling module is to assign a label for each word in the sentence to encode the boundaries of the terms and definitions (if any). In particular, for each word $w_i \in W$, we use the concatenation $h'_i$ of its BiLSTM output vector $h_i$ and GCN output vector $\hat{h}_i$ (i.e., $h'_i = [h_i, \hat{h}_i]$) as the feature vector to perform the label prediction for $w_i$. Essentially, based on the feature vector $h'_i$, we first transform it into a score vector $S_i$ whose dimensions correspond to the possible word labels/tags (i.e., the five BIO tags) and quantify the possibility for $w_i$ to receive the corresponding labels: $S_i = W_S h'_i$ where $W_S$ is the trainable weight matrix and $|S_i| = 5$.


In the next step, we feed the score vectors $S_i$ for the words $w_i$ into a conditional random field (CRF) layer to compute the scores to quantify the possibilities of the possible label sequences $\hat{l}_1, \hat{l}_2, \ldots, \hat{l}_N$ for the words in $W$. The purpose of this CRF layer is to capture the dependencies between the BIO labels/tags that have been shown to be useful for the other sequence labeling tasks in NLP \cite{Lafferty:01}. In particular, the score for a possible label sequence $\hat{l}_1, \hat{l}_2, \ldots, \hat{l}_N$ for $W$ would be:
\begin{equation}
    Score(\hat{l}_1,\hat{l}_2,...,\hat{l}_N|W) = \Sigma_{j=1}^N \big( S_{\hat{l}_j} + T_{\hat{l}_{j-1},\hat{l}_{j}} \big)
\end{equation}

where $T$ is the trainable transition matrix for the BIO labels. For CRF, we compute the normalization score to form the probability distribution $P_{labeling}(\hat{l}_1,\hat{l}_2,...,\hat{l}_N|W)$ for the possible label sequences for $W$ via dynamic programming \cite{Lafferty:01}. We use the negative log-likelihood $L_{labeling}$ of the input example as the objective function for this sequence labeling module:
\begin{equation}
    L_{labeling} = -\log P_{labeling}(l_1,l_2,\ldots,l_N|W)
\end{equation}
where $L = l_1,l_2,\ldots,l_N$ is the golden label sequence for $W$. Finally, the Viterbi decoder is employed to infer the sequence of labels with highest score for the input sentence.

\subsubsection{Sequence Classification}


In this module, we need to predict the label for the input sentence $W$ to indicate whether $W$ contains a pair of a term and definition or not (i.e., a binary decision). For this task, we first obtain a representation vector $\hat{h}^S$ for the input sentence by aggregating the syntax-enriched and context-aware vectors $\hat{H}=\hat{h}_1,\hat{h}_2,\ldots,\hat{h}_N$ from GCN with the max-pooling operation: $\hat{h}^S = Max\_Pooling(\hat{h}_1,\hat{h}_2,\ldots,\hat{h}_N)$.

As the vectors in $\hat{H}$ are obtained via GCN over the whole dependency tree of $W$ and BiLSTM vectors, we expect that their aggregated vector $\hat{h}^S$ would be able to capture the most important context features for the sequence classification task via max pooling. Consequently, $\hat{h}^S$ would be fed into a 2-layer feed forward neural network with a softmax layer in the end to compute the probability distribution $P_{classification}(.|W)$ over the two possibilities for the label of $W$ (i.e., definitional or not). This probability distribution would be used for both prediction and training. We use also negative log likelihood as the loss function $L_{classification}$ for the sequence classification module in this work:
\begin{equation}
    L_{classification} = -\log P_{classification}(l|W)
    \label{eq:sc}
\end{equation}
where $l$ is the true definitional label for $W$.




\subsubsection{Syntactic Connection between Term and Definition}


The GCN module in the encoding is one way to exploit the whole dependency structure of the input sentence to infer effective representations for the labeling and classification tasks. However, the application of GCN for the dependency structure is agnostic to the positions of the terms and definitions as well as the relevant context words to connect the terms and the definitions in the sentences. In this work, we argue that if the representation vectors from the GCN module are trained to be more aware of the terms and the definitions in the sentences, they can be more customized for the DE tasks and achieve better performance. Motivated by this idea, from the dependency tree and the term and definition in $W$, we identify the words along the shortest dependency path $SP$ between the term and the definition in the sentence. This information allows us to assign a binary label $d_i$ for each word $w_i \in W$ where $d_i = 1$ if $w_i$ belong to the dependency path and 0 otherwise. Afterward, we encourage the GCN vectors $\hat{H} = \hat{h}_1,\hat{h}_2,\ldots,\hat{h}_N$ to be aware of the terms and definitions by using the vectors $\hat{h}_i$ to predict the membership on the dependency path $SP$ of $w_i$. In particular, the GCN vectors $\hat{h}_i$ would be consumed by a 2-layer feed forward neural network with a softmax in the end to obtain the distribution $P^{dep}_i(.|W)$ over the two possibilities of $w_i$ to belong to $SP$ or not. We would then optimize the negative log-likelihood $L_{dep}$ for the dependency path prediction task based on the distributions $P^{dep}_i(.|W)$ and the ground-truth sequence label $D = d_1, d_2, \ldots, d_N$ for the words in $W$:
\begin{equation}
    L_{dep} = -\sum^N_{i=1} \log P^{dep}_i(d_i|W)
    \label{eq:dp}
\end{equation}

Note that as the term and definition in $W$ might have multiple words, we use the lowest node among the common ancestors of the pairs of words in the term and definition to determine the dependency path $SP$. For the sentences that do not contain any pair of terms and definitions, we simply set $d_i = 0$ for every word in the sentences.

\subsubsection{Semantic Consistency}

As presented in the introduction, we seek to enforce the semantic consistency between the representations of the terms, definitions, and the entire sentences to improve the representations induced by the model for DE. In this work, we achieve this goal at both the local and global levels.

\textbf{The Local Level}

At the local level, we aim to ensure that the representations of the terms and the definitions are similar to each other due to their reference to the same concepts in practice. In order to implement this idea, we first compute the representations for the term and definition in $W$ using the max pooling operation. In particular, let $start_T$ and $end_T$ be the indexes of the first and last token for the term in $W$ respectively (i.e., $1 \le start_T \le end_T \le N$). Similarly, let $start_D$ and $end_D$ denote the starting and ending indexes of the words in the definition in $W$ (if any). The representations $h^T$ and $h^D$ for the term and definition in $W$ are then computed by: $h^T = Max\_Pooling(h_i | start_T \le i \le end_T)$ and  $h^D = Max\_Pooling(h_i | start_D \le i \le end_D)$.


Note that we use the BiLSTM output vectors $h_i$ instead of the GCN vectors $\hat{h}_i$ to compute semantic representations for the terms and representations in this case as we would like to avoid the confusion between semantics and syntax for the GCN vectors $\hat{H}$ that have been intended to encode rich syntactic context information for $W$.

As the terms and the definitions are naturally related, the semantic consistency between $h^T$ and $h^D$ would be extensively exploited in this work. In particular, we would enforce the similarity between $h^T$ and $h^D$ both directly and indirectly in this work. First, for the direct enforcement, we attempt to maximize the dot product between the two vectors in the loss function:
\begin{equation}
    L^1_{sem} = - h^T h^D
    \label{dlsc}
\end{equation}
Second, in the indirect enforcement, we seek to improve the consistency/similarity of $h^T$ and $h^D$ by ensuring that $h^D$ is more similar to the representation vectors for the words in the term (i.e., $h_i$ for $start_T \le i \le end_T$) than the vectors for the words in the sentence (i.e., $h_i$ for $i \not\in I = [start_T, end_T] \cup [start_D, end_D]$). To this goal, we employ a discriminator $DI$ that can produce a similarity score between $h^D$ and the representation vectors for the other words in the sentence. In particular, for some word $w_i$, the input for $DI$ is the concatenation of its BiLSTM representation vector $h_i$ and the definition representation $h^D$ (i.e., $[h_i,h^D]$). $DI$ would feed $[h_i, h^D]$ into a 2-layer feed forward neural network that uses the sigmoid activation function $\sigma$ and produces a single scalar $\hat{s}_i$ to represent the similarity of $h_i$ and $h^D$ ($0 \le \hat{s}_i \le 1$): $\hat{s}_i = DI(h_i, h^D)$. Afterward, in the training process, the model would be tasked to increase the similarity scores between $h^D$ and the representation vectors for the words in the term, and decrease the similarity between $h^D$ and the vectors for the other words. The loss function in this case would be:
\begin{equation}
\small
    L^2_{sem} = \sum^{end_T}_{i=start_T} \log DI(\hat{s}_i) + \sum_{i \not\in I} \log (1 - 
    DI(\hat{s}_i))
    \label{ilsc}
\end{equation}


\textbf{The Global Level}

At the global level, we attempt to enforce the consistency between the representation vectors for the whole sentence and the term-definition pair in $W$. The goal is to encourage the representation vector for the sentence $W$ to mainly focus on the information about the term and definition presented in the sentence, thereby reducing the effect of the irrelevant words in the representation vector of $W$ for the DE problems. Similar to the local level, we also start by computing the representation vectors $h^S$ and $h^{TD}$ for the sentence and the term-definition pair respectively via the max pooling operation: $h^S = Max\_Pooling(h_1,h_2,\ldots,h_N)$ and $h^{TD} = Max\_Pooling(h_i | i \in I)$.


We can also follow the direct approach in the local level to promote the consistency between $h^S$ and $h^{TD}$. However, as $h^S$ might need to involve the information for the important context words between the term and definition in $W$ (in addition to the information about the term and the definition themselves), the direct consistency with the dot product might be too strict that can eliminate the important information about the context. Consequently, in this global level, we only aim to introduce a mild and indirect consistency constraint for $h^S$ and $h^{TD}$. In particular, in a latent label prediction framework, we ensure that $h^S$ can be used to predict the same latent label as the one predicted by $h^{TD}$, implicitly requiring $h^S$ to maintain the most important information in $h^{TD}$. In order to implement this idea, we select a fixed number $U$ of latent labels. Afterward, we feed $h^S$ and $h^{TD}$ into a feed forward neural network with the softmax layer in the end to obtain the probability distributions $P^{sem}_S(.|W)$ and $P^{sem}_{TD}(.|W)$ (respectively) over the $U$ latent labels. We would then obtain the latent label $l_{TD}$ predicted by $h^{TD}$ via the argmax function: $l_{TD} = argmax_y P^{sem}_{TD}(y|W)$.

In the next step, $l_{TD}$ would be used as the ground-truth latent label to compute the negative log-likelihood $L^3_{sem}$ based on the $P^{sem}_S(.|W)$ distribution that would be optimized in the training process:
\begin{equation}
    L^3_{sem} = -\log P^{sem}_S(l_{TD}|W)
    \label{gsc}
\end{equation}
To summarize, the total loss for the semantic consistency for DE would be: $L_{sem} = a L^1_{sem} + b L^2_{sem} + c L^3_{sem}$.

Finally, the overall loss to train the model in this work would be: $L_{all} = \alpha L_{labeling} + \beta L_{classification} +  \gamma L_{dep} + \eta L_{sem}$
where $a$, $b$, $c$, $\alpha$, $\beta$, $\gamma$, and $\eta$ are the trade-off parameters to be tuned on the development datasets.

\section{Experiments}

\begin{table*}[t!]
 \small
\begin{center}
 \resizebox{.85\textwidth}{!}{
\begin{tabular}{l|ccc|ccc|ccc|ccc}
  & \multicolumn{3}{c}{WCL} & \multicolumn{3}{c}{W00} & \multicolumn{3}{c}{Textbook} & \multicolumn{3}{c}{Contract} \\
  & P & R & F1 & P & R & F1 & P & R & F1 & P & R & F1 \\
  \hline
    DefMiner  \cite{jin:2013} & 82.0 & 78.5 & 80.5 & 52.5 & 49.5 & 50.5 & - & - & - & - & - & - \\
    LSTM-CRF \cite{li2016definition} & 83.4 & 80.5 & 81.7 & 57.1 & 55.9 & 56.2 & 46.7 & 47.5 & 47.0 & 63.2 & 73.5 & 67.4 \\
    GCDT \cite{Liu:19} & 82.5 & 81.3 & 81.2 & 57.9 & 56.6 & 57.4 & 45.8 & 48.9 & 47.0 & 63.5 & 73.2 & 66.7 \\
    CVT \cite{clark2018semi} & 83.0 & 82.3 & 82.5 & 59.3 & 57.9 & 58.9 & 46.1 & 49.2 & 47.3 & 64.1 & 74.5 & 68.2 \\
    \textbf{Ours} & 85.1 & 81.9 & 83.3 & 60.9 & 60.3 & 60.6 & 52.8 & 48.7 & 50.6 & \textbf{66.1} & \textbf{76.1} & \textbf{71.7} \\
    \hline
        \textbf{Ours + BERT} & \textbf{87.1} & \textbf{83.8} & \textbf{85.3} & \textbf{66.9} & \textbf{67.3} & \textbf{66.9} & \textbf{53.8} & \textbf{54.5} & \textbf{54.0} & 65.7 & 74.3 & 68.2 \\
\end{tabular}
 }
\end{center}
\caption{\label{word-perf} Sentence Labeling Performance.  The WCL and W00 results are based on the 10-fold cross validation performance while the results for DEFT (i.e., Textbook and Contract) are obtained on the test sets. The proposed model is significantly better the baselines ($p < 0.01).$
  }
\end{table*}

\subsection{Dataset \& Hyper Parameters}
We evaluate our model on three benchmark datasets for DE:

$\bullet$\textbf{WCL}: Word Class Lattices (WCL) was introduced by \cite{navigli2010learning}. It consists of 1,871 definitional and 2,847 non-definitional sentences from Wikipedia. The term and definitions of WCL belong to the general domain.
    
$\bullet$\textbf{W00}: This dataset is contributed by \cite{jin:2013}. It has 731 definitional and 1454 non-definitional sentences from the ACL-ARC anthology. The definitions in W00 are from the scientific domain (i.e., the NLP papers).

    
$\bullet$\textbf{DEFT}: This is a recently released dataset for DE \cite{spala2019deft}. DEFT consists of two categories of definitions: a) Contracts: involving 2,433 sentences from the 2017 SEC contract filing with 537 definitional and 1906 non-definitional sentences. Besides terms and definitions, this corpus has an additional type \textit{qualifier}. It indicates the words/phrases specifying the conditions, dates or locations in which the definitions are valid for the terms. We also use the BIO tagging schema for this type. 2) Textbook: involving 21,303 sentences from the publicly available textbooks in different domains, including biology, history, and physics. This corpus contains 5,964 definitional and 15,339 non-definitional sentences.

For all the datasets, we use the standard data splits to ensure a comparable comparison with the prior work. We fine tune the model parameters on the validation set of the DEFT Contract dataset and fix the detected parameters to train and evaluate the models on the other datasets for consistency. The parameters we found include: 50 dimensions for the POS embeddings; 200 dimensions for the LSTM and GCN hidden vectors and all the feed forward neural networks in the model; $a=1,b=1,c=1,\alpha=1, \beta=10, \gamma=1$ and $\eta=1$ for the trade-off parameters; $U=3$ for the latent labels in the semantic consistency module; and the learning rate of 0.003 for the Adam optimizer. We use the pre-trained word embeddings {\tt GloVe} with 300 dimensions from \cite{pennington2014glove} to initialize the model. Finally, to assess how well our model could benefit from the pre-trained contextualized word embeddings, we also perform an additional experiment where BERT \cite{Devlin:18} is used to initialize the word embeddings in the model.


\subsection{Results}
\subsubsection{Sequence Labeling Performance}

This section evaluates the models on the sequence labeling task for DE. As the evaluation metrics, following previous work on word-level DE, we use macro averaged Precision, Recall and F1 score of all distinct classes (i.e., Term and Definition in WCL and W00, and Term, Definition and Qualifier in DEFT). We compare the proposed model with the following baselines:

\textbf{DefMiner}: A feature engineering model for DE, using 12 hand crafted linguistic features in \cite{jin:2013}. DefMiner is the state-of-the-art feature-based model for sequence labeling DE on the WCL and W00 datasets.

\textbf{LSTM-CRF} A deep learning model for sequence labeling for DE based on LSTM and CRF \cite{li2016definition}.

\textbf{GCDT}: A very recent deep learning method for sequence labeling \cite{Liu:19}. GCDT enriches the pre-trained word embeddings with the representations of the sentence and the character-based word embeddings for an LSTM-based decoder.
    

\textbf{CVT}: A recent state-of-the-art deep learning model for the general sequence labeling problem that enriches the partial representations of the input sentences with the prediction from the full representation \cite{clark2018semi}. Essentially, CVT employs the predictions obtained from the full representation of the input as the golden labels to train auxiliary models to exploit partial representations of the input (e.g., only the forward LSTM instead of BiLSTM). We use the released implementation of CVT to evaluate it on the DE datasets in this work for a fair comparison.

Table \ref{word-perf} shows the performance of the models on all the four datasets
The most important observation from the table is that the proposed model significantly outperform the baseline models across different matrices (only except the recall in the WCL dataset). The performance gap is substantial on the Textbook and Contract datasets (i.e., up to 3.3\% improvement on the absolute F1 score over the second best system CVT), clearly demonstrating the effectiveness of the proposed model in this work. This table also shows that our model can benefit from the contextualized embeddings (e.g., BERT \cite{Devlin:18}) as BERT can significantly improve the proposed model over three out of four datasets.

\subsubsection{Sequence Classification Performance}

This section evaluates the models on the sentence classification task for DE. Due to its popularity in the DE literature for this setting, we first report the performance of the models on the general domain dataset WCL. The following baselines are chosen for comparison in this dataset:


\textbf{Feature engineering models}: These models perform the classification using the hand designed features from the input text (i.e., WCL \cite{navigli:2010} and DefMiner \cite{jin:2013}).

\textbf{Structure based models}: These models benefit from the structure information in the given sentence to perform classification (i.e., B\&DC \cite{boella2014learning} and E\&S \cite{espinosa2016defext}).

\textbf{Deep learning models}: These models employ deep learning architectures (e.g., LSTM and CNN) for classification (i.e., LSTM-POS \cite{li2016definition} and SA \cite{anke:2018}). SA is the state-of-the-art model for sequence classification DE on WCL and W00 datasets.
    

The performance of the models is shown in Table \ref{sent-perf}. It is clear from the table that the deep learning models are generally better than the feature engineering and structure-based models with large performance gap. Among the deep learning models, the proposed model significantly outperforms the other models ($p < 0.01)$ (i.e., up to 5\% improvement on the absolute F1 score over SA), thereby further confirming the advantage of the proposed model for DE in this work.


\begin{table}[t!]
 \small
\begin{center}
 \resizebox{.44\textwidth}{!}{
\begin{tabular}{l|ccc}
  & P & R & F1 \\
  \hline
    WCL \cite{navigli:2010} & 98.8 & 60.7 & 75.2 \\
    DefMiner \cite{jin:2013} & 92.0 & 79.0 & 85.0 \\
    \hline
    B\&DC \cite{boella2014learning} & 88.0 & 76.0 & 81.6 \\
    E\&S \cite{espinosa2016defext} & 85.9 & 85.3 & 85.4 \\
    \hline
    LSTM-POS \cite{li2016definition} & 90.4 & 92.0 & 91.2 \\
    SA \cite{anke:2018} & 94.2 & 94.2 & 94.2 \\
  \hline
    \textbf{Ours} & \textbf{99.7} & \textbf{99.4} & \textbf{99.5} \\
\end{tabular}
 }
\end{center}
\caption{\label{sent-perf} Sequence classification performance on WCL. The results are based on the 10-fold cross validation with the same data splits.}
\end{table}

\begin{table}[t!]
 \small
\begin{center}
 \resizebox{.4\textwidth}{!}{
\begin{tabular}{l|ccc|ccc}
  & \multicolumn{3}{c}{Textbook} & \multicolumn{3}{c}{Contract} \\
  & P & R & F1 & P & R & F1 \\
  \hline
    Full Model &  52.8 & 48.7 & 50.6 & 66.1 & 76.1 & 71.7  \\
    Full - SC & 50.5 & 47.8 & 48.8 & 65.0 & 73.8 & 68.5 \\
    Full - GCN & 51.1 & 47.3 & 48.8 & 65.0 & 74.3 & 68.5 \\
    Full - DPP & 50.3 & 46.9 & 48.1 & 64.2 & 74.2 & 68.2 \\
    Full - DLSC & 50.9 & 47.2 & 49.6 & 66.2 & 75.6 & 70.5 \\
    Full - ILSC & 50.9 & 47.8 & 49.8 & 65.8 & 74.1 & 69.6 \\
    Full - GSC & 51.2 & 47.5 & 49.0 & 65.2 & 74.5 & 69.1 \\
    
\end{tabular}
 }
\end{center}
\caption{\label{subtask-ablation} Sequence labeling performance for the ablation study.
}


\end{table}

Regarding the W00 and DEFT datasets for the sequence classification setting, we compare the proposed model with the state-of-the-art SA model in \cite{anke:2018}
The models' performance is presented in Table \ref{sent-perf-DEFT}, clearly showing that the proposed model is significantly superior to SA over different metrics and datasets.

\begin{table}[t!]
\begin{center}
 \resizebox{.48\textwidth}{!}{
\begin{tabular}{l|ccc|ccc|ccc}
  & \multicolumn{3}{c}{W00} & \multicolumn{3}{c}{Textbook} & \multicolumn{3}{c}{Contract} \\
  & P & R & F1 & P & R & F1 & P & R & F1 \\
  \hline
    SA & 52.0 & 67.6 & 57.4 & 70.1 & 57.8 & 64.2 & 83.3 & 84.9 & 84.1 \\
    \textbf{Ours} & \textbf{67.0} & \textbf{68.0} & \textbf{67.2} & \textbf{75.0} & \textbf{66.1} & \textbf{70.3} & \textbf{88.1} & \textbf{95.6} & \textbf{91.7}
\end{tabular}
 }
\end{center}
\caption{\label{sent-perf-DEFT} The sequence classification performance of the models on W00, Textbook, and Contract.
}
\end{table}

\subsection{Analysis}
\subsubsection{Ablation Study}

We have seven major components of the model in this work, i.e., sequence labeling, sequence classification ({\bf SC}) (i.e., Equation \ref{eq:sc}), the graph convolutional neural networks ({\bf GCN}), dependency path prediction ({\bf DPP}) for the syntactic connection (i.e., Equation \ref{eq:dp}), the direct and local semantic consistency ({\bf DLSC}) (i.e., Equation \ref{dlsc}), the indirect and local semantic consistency ({\bf ILSC}) (i.e., Equation \ref{ilsc}), and the global semantic consistency ({\bf GSC}) (i.e., Equation \ref{gsc}). Considering sequence labeling as the main DE task (due to its more challenging nature for DE)
, we exclude the other six components from the proposed model one by one to evaluate their contribution to the overall model. Table \ref{subtask-ablation} presents the performance of the models on the test sets of Textbook and Contract, the two largest datasets in this work.

As we can see from the table, all the proposed components in the model are necessary for the sequence labeling setting of DE as removing any of them would hurt the model's performance significantly on both Textbook and Contract. Specifically, the removal of the sequence classification module SC would reduce the F1 scores by 1.8\% and 3.2\% on the Textbook and Contract datasets respectively, thus demonstrating the benefit of the jointly inference for sequence labeling and sequence classification for DE in this work. Among the semantic consistency constraints, the global consistency (i.e., GSC) seems more important than the local consistency (i.e., DLSC and ILSC) due to the larger performance reduction caused by GSC.

Interestingly, the dependency path prediction (i.e., DPP) seems to contribute the most to the performance of the proposed model as its absence would lead to the largest performance loss of the model (i.e., a F1 reduction of 2.5\% on Textbook and 3.5\% on Contract). The goal of DPP is to capture the important context words between the terms and definitions that might be far away from each other via the dependency structures. As the GCN also relies on the dependency structures to model the syntactically neighboring context words at each layer, a natural question is whether we can replace DPP with a deeper GCN model (i.e., more layers) to increase the dependency coverage (i.e., more hops) for each word and implicitly encode the dependency paths between the terms and definitions. Consequently, we evaluate the performance of the proposed model when DPP is eliminated, but more layers of GCN are applied in Table \ref{ablation-gcn} (on the Textbook and Contract test sets). As we can see from the tables, when the DPP component is not incorporated, simply increasing the depth of the GCN module is not sufficient to model the dependency paths between the terms and definitions (i.e., the performance of the model in such cases are significantly worse than those of the initially proposed model). This clearly demonstrates the benefits of the dependency path prediction proposed in this work for DE.

\begin{table}[t!]
\begin{center}
\resizebox{.48\textwidth}{!}{
\begin{tabular}{l|ccc|ccc}
  & \multicolumn{3}{c}{Textbook} & \multicolumn{3}{c}{Contract} \\
  & P & R & F1 & P & R & F1 \\
  \hline
    Full Model & 52.8 & 48.7 & 50.6 & 66.1 & 76.1 & 71.7  \\
    Full - DPP (2-layer GCN) & 51.1 & 47.3 & 48.8 & 65.0 & 74.3 & 68.5 \\
    Full - DPP (3-layer GCN) & 50.4 & 47.1 & 48.5 & 64.8 & 74.1 & 68.0 \\
    Full - DPP (4-layer GCN) & 49.2 & 46.9 & 48.0 & 64.6 & 73.8 & 67.3 \\
\end{tabular}
}
\end{center}
\caption{\label{ablation-gcn} The model's performance with different numbers of layers for GCN when DPP is excluded.
  }
\end{table}

\subsubsection{Per Class Performance}

We study the per class performance of the proposed model for terms and definitions in this section. In particular, Table \ref{perclass} reports the precision, recall and F1 score per class (i.e, B-Term, I-Term, B-Definition, and I-Definition) of the model on the Textbook and Contract test sets. In general, we see that extracting terms is more manageable than definitions on both Textbook and Contract due to the better performance of B-Term over B-Definition and I-Term over I-Definition (except for I-Term vs I-Definition on Textbook). This is reasonable as definitions are often presented in much more complicated expressions than terms. 




\begin{table}[t!]
 \small
\begin{center}
\resizebox{.4\textwidth}{!}{
\begin{tabular}{lcccccc}
  & \multicolumn{3}{c}{Textbook} & \multicolumn{3}{c}{Contract} \\
  & P & R & F1 & P & R & F1 \\
  \hline
    B-Term & 68.1 & 57.5 & 62.4 & 84.0 & 93.2 & 88.4  \\
    I-Term & 60.6 & 52.3 & 56.1 & 85.5 & 94.3 & 89.6 \\
    B-Definition & 62.8 & 57.0 & 59.8 & 62.8 & 67.4 & 65.0 \\
    I-Definition & 65.6 & 60.6 & 63.0 & 78.5 & 88.4 & 83.2 \\
\end{tabular}
}
\end{center}
\caption{\label{perclass} Per class performance on Textbook and Contract datasets of DEFT corpora.
  }
\end{table}

\section{Conclusion}

We introduce a novel model for the problem of definition extraction that presents a multi-task learning framework to jointly perform sequence classification and sequence labeling for this problem based on deep learning. In order to improve the representations learned by the model, we propose several mechanisms to exploit the whole dependence structures of the input sentences and the semantic consistency between the terms, the definitions and the sentences. 
We achieve the state-of-the-art performance on four benchmark datasets. 


\section*{Acknowledgments}

This work has been supported in part by Vingroup Innovation Foundation (VINIF) in project code VINIF.2019.DA18, IARPA  BETTER, the NSF grant CNS-1747798 to the IUCRC Center for Big Learning, and Adobe Research Gift.

\bibliographystyle{aaai}
\bibliography{main.bib}

\end{document}